\documentclass[acmlarge]{acmart}

\usepackage{booktabs} 
\usepackage{relsize}

\usepackage[ruled]{algorithm2e} 

\SetAlFnt{\small}
\SetAlCapFnt{\small}
\SetAlCapNameFnt{\small}
\SetAlCapHSkip{0pt}
\IncMargin{-\parindent}

\acmJournal{JDIQ}
\acmVolume{9}
\acmNumber{4}
\acmArticle{39}
\acmYear{2019}
\acmMonth{4}
\acmArticleSeq{11}

\setcopyright{usgovmixed}

\acmDOI{0000001.0000001}

\received{March 2018}
\received{March 2018}
\received[accepted]{July 2018}

\begin{document}
\title[Combining Similarity Features and Deep Representation Learning for Automatic Stance Detection]{Combining Similarity Features and Deep Representation Learning for Stance Detection in the Context of Checking Fake News}


\author{Lu\'is Borges}
\affiliation{%
  \institution{INESC-ID, Instituto Superior T\'ecnico, Universidade de Lisboa}
  \streetaddress{Rua Alves Redol, 9}
  \city{Lisbon}
  \postcode{1000-029}
  \country{Portugal}}
\email{luis.borges@tecnico.ulisboa.pt}
\author{Bruno Martins}
\affiliation{%
  \institution{INESC-ID, Instituto Superior T\'ecnico, Universidade de Lisboa}
  \streetaddress{Rua Alves Redol, 9}
  \city{Lisbon}
  \postcode{1000-029}
  \country{Portugal}
}
\email{bruno.g.martins@tecnico.ulisboa.pt}
\author{P\'avel Calado}
\affiliation{%
  \institution{INESC-ID, Instituto Superior T\'ecnico, Universidade de Lisboa}
  \streetaddress{Rua Alves Redol, 9}
  \city{Lisbon}
  \postcode{1000-029}
  \country{Portugal}}
\email{pavel.calado@tecnico.ulisboa.pt}

\begin{abstract}
Fake news are nowadays an issue of pressing concern, given their recent rise as a potential threat to high-quality journalism and well-informed public discourse. The Fake News Challenge (FNC-1) was organized in early 2017 to encourage the development of machine learning-based classification systems for stance detection (i.e., for identifying whether a particular news article agrees, disagrees, discusses, or is unrelated to a particular news headline), thus helping in the detection and analysis of possible instances of fake news. This article presents a novel approach to tackle this stance detection problem, based on the combination of string similarity features with a deep neural network architecture that leverages ideas previously advanced in the context of learning efficient text representations, document classification, and natural language inference. Specifically, we use bi-directional Recurrent Neural Networks (RNNs), together with max-pooling over the temporal/sequential dimension and neural attention, for representing (i) the headline, (ii) the first two sentences of the news article, and (iii) the entire news article. These representations are then combined/compared, complemented with similarity features inspired on other FNC-1 approaches, and passed to a final layer that predicts the stance of the article towards the headline. We also explore the use of external sources of information, specifically large datasets of sentence pairs originally proposed for training and evaluating natural language inference methods, in order to pre-train specific components of the neural network architecture (e.g., the RNNs used for encoding sentences). The obtained results attest to the effectiveness of the proposed ideas and show that our model, particularly when considering pre-training and the combination of neural representations together with similarity features, slightly outperforms the previous state-of-the-art.
\end{abstract}

%
%


%
%



\maketitle

\renewcommand{\shortauthors}{Borges et al.}

\section{Introduction}

Fake news (i.e., made up stories with the intention of deceiving, and that most of the times are used to achieve secondary gains) are undoubtedly one of the most serious challenges that journalists and the news industry are facing today. Given the ease in obtaining and spreading information through social networking platforms, it is increasingly harder to know for sure what to trust, with the absorption of fake news by the masses having increasingly harmful consequences~\cite{vosoughi2018spread}. Automatically dealing with fake news has drawn considerable attention in several research communities~\cite{lazer2018science,perez2017automatic,shu2017exploiting,shu2018fakenewsnet,tosikdebunking,DeClarE2018,konstantinovskiy2018towards}. However, the task of evaluating the veracity of news articles is still very demanding and complex, even for trained specialists and much more for automated systems.

A useful first step towards identifying fake news articles relates to understanding
what other news agencies, in a given moment, are reporting about the same topic. This sub-task is often referred to as stance detection, and automating this process might be useful in developing automated assistants to help in fact checking. In particular, an automatic approach to stance detection would allow, for example, someone to insert an allegation or a news title, and recover related articles that either agree, disagree, or discuss that title. Then, the human checker would use her own judgment to assess the situation. 

Based on the aforementioned general ideas, a Fake News Challenge (FNC-1) was organized in early 2017 by a group of academics and contributors from the journalism industry to foster the development of systems applying artificial intelligence and machine learning for evaluating what a news source is saying about a particular issue. Specifically, FNC-1 involved developing models that, given the body of a news article and a news headline, estimate the stance of the article towards the headline (i.e., the article can either agree, disagree, discuss, or be unrelated to the headline). More information on the FNC-1 task, its rules, the dataset, and the evaluation metrics, can be found on the official website\footnote{\scriptsize{\url{http://fakenewschallenge.org}}}. A total of 50 teams actively participated on the challenge, and the training/testing data splits have been released afterwards, in order to encourage further developments. 

The winning entries in FNC-1 relied on ensemble models combining similarity features computed between the headline and the body, with representations built from the words occurring in the texts (e.g., by processing word embeddings with convolutional neural networks). Despite the good results achieved by the participants, with scores of 82.02\% and 81.98\% for the two winning teams, in terms of the accuracy metric considered for the challenge, we believe there are several opportunities for further improvements. For instance, although deep neural networks are known to outperform other approaches (e.g., linear models based on careful feature engineering) in several natural language processing problems related to stance detection (e.g., in tasks related to measuring semantic similarity between sentences, and/or related to performing natural language inference), the results from FNC-1 showed that standard approaches based on convolutional and/or recurrent neural networks are not well-suited to this task, as they fail to model semantic relations with respect to large pieces of text (i.e., the bodies of news articles, composed of multiple sentences). A FNC-1 baseline that was introduced by the task organizers, which leverages a combination of feature engineering heuristics (i.e., word and $n$-gram overlap features, as well as indicators for refutation and polarity), achieves a competitive performance with the best systems, and it even outperforms several widely used deep learning architectures. 

In this article, we address the stance detection problem proposed in the context of FNC-1, using a novel approach based on the combination of similarity features with deep neural networks for generating effective representations. The main ideas and contributions of our work can be summarized as follows:

\begin{itemize}
\item Taking inspiration on previous studies addressing natural language inference~\cite{snli,inference}, we propose a deep neural network architecture for stance detection, as defined in the FNC-1 task. Our architecture leverages pre-trained word embeddings, uses shared parameters for some of the components (e.g., the sentence encoders), and attempts to match multiple representations learned from the inputs. Specifically, we combine/match representations inferred from three different inputs (i.e., the headline, the first two sentences of the news article, and the entire document) through a series of operations known to perform well on natural language inference~\cite{supervised}, namely the vector concatenation, difference, and element-wise product. A final layer processes this result, and predicts the stance of the news article towards the headline.

\item The proposed neural network architecture leverages a hierarchical approach for modeling the body of news articles, taking inspiration on previous studies addressing the classification of long documents~\cite{hierarchical,franciscoduarte}. In this approach, a Recurrent Neural Network (RNN) is used for modeling the sequence of sentences, which in turn are individually modeled by a separate RNN encoding sequences of words. Specifically, we evaluated the use of bi-directional RNNs (i.e., either Gated Recurrent Units or Long Short-Term Memory units, optionally also in a stacked arrangement with shortcut connections), together with max-pooling and/or a neural attention mechanism that weights the individual word representations, for encoding sentences in our model (i.e., the headline, and the sentences in the news article being matched). Separate layers of bi-directional RNNs, also combined with max-pooling and/or neural attention, are used for encoding the sentences that constitute the document. This approach can directly incorporate knowledge of the document structure in the model architecture, at the same time also exploring the intuition that not all sentences/words will be equally relevant for predicting the stance;

\item We used the SNLI~\cite{snli} and MultiNLI~\cite{inference} datasets of sentence pairs, previously proposed for evaluating natural language inference models (i.e., models for checking if a given hypothesis sentence entails, contradicts, or is neutral towards a given premise sentence), to pre-train some of the components involved in our neural network architecture. Previous studies have already attested to the benefits of similar pre-training procedures in other natural language processing and text classification tasks~\cite{supervised}. We specifically used SNLI/MultiNLI sentence pairs to pre-train the components involved in encoding sentences, and also the part of the network that matches the headline with the first two sentences of the document (although this second component is not used in the neural architecture that achieved the best experimental results);

\item Learning effective representations for news article bodies can be quite challenging for neural methods, although this is essential for achieving good performance on the FNC-1 task~\cite{combination,Mohtarami2018}. We therefore propose to combine the representations learned through neural network layers, with external similarity features. We specifically leverage features from previous FNC-1 methods (e.g., from the baseline method introduced by the organizers of the challenge), together with other methods proposed for similar text matching problems (e.g., BLEU and ROUGE scores~\cite{bleu,rouge}, the soft-cosine similarity metric~\cite{charlet2017simbow}, or the CIDEr score~\cite{cider}).

\item We report on the results of an extensive set of experiments, evaluating the contribution of each of the aforementioned components. The results confirm that model pre-training can indeed improve the overall accuracy. Moreover, despite the use of the hierarchical attention method for encoding the news articles, which has been shown to perform well on tasks related to the classification of long documents~\cite{hierarchical}, our results also show that hand-crafted similarity features are highly beneficial. The complete method establishes a new state-of-the-art result for the FNC-1 dataset, slightly outperforming the previous approach described by Bhatt et al.~\cite{combination}. The source code supporting the experiments reported on this article has also been made available online through GitHub\footnote{\scriptsize{\url{http://github.com/LuisPB7/fnc-msc}}}.
\end{itemize}

The remainder of this article is organized as follows: Section 2 presents fundamental concepts related to the use of deep neural networks in tasks such as natural language inference or stance detection, together with an overview of previous work related to FNC-1. Section 3 presents the neural network architecture proposed for handling the FNC-1 stance detection task, detailing (i) the use of bi-directional RNNs together with max-pooling and with an attention mechanism for encoding sentences, (ii) the hierarchical attention model for encoding larger pieces of text (i.e., the body of a news article), (iii) the method used for combining the representations generated for the headline and the body, (iv) the integration of similarity features inspired on other approaches for the FNC-1 task, and (v) the use of the SNLI and MultiNLI datasets for model pre-training. Section 4 describes the experimental evaluation of the proposed method, specifically detailing the evaluation methodology and presenting ablation tests that validate the contribution of the different model components. Finally, Section 5 presents the main conclusions and highlights possible directions for future work.

\section{Concepts and Related Work}

This section starts by reviewing neural network methods for modeling textual information, which are the fundamental building blocks of most modern approaches for tasks such as natural language inference or stance detection. It then overviews previous work developed in the context of the Fake News Challenge (FNC-1).

\subsection{Deep Neural Networks for Natural Language Processing}

Supervised machine learning is extensively used for Natural Language Processing (NLP). In general, supervised learning concerns with inferring the parameters of models that take as input vector representations $\boldsymbol{x}$ and return as output another vector, where each dimension reflects the probability of the input belonging to a certain class. In NLP tasks such as document classification, $\boldsymbol{x}$ typically encodes features like words or characters occurring in the text. Bag-of-words approaches, and extensions considering $n$-grams, are arguably the most commonly used representations, treating words and/or phrases as unique discrete symbols, and weighting their contributions through heuristics such as the Term Frequency multiplied by the Inverse Document Frequency (TF-IDF). More recently, noting that bags-of-words often fail in capturing similarities between words, at the same time suffering from sparsity and high dimensionality, methods using neural networks to learn distributed vector representations of words (i.e., word embeddings) have gained popularity. These word embeddings can be (pre-)trained in an unsupervised manner over large corpora (e.g., by learning to predict target words using their neighboring words), through methods like \textit{word2vec}~\cite{word2vec} or GloVe~\cite{glove}. One can then average the embedding vectors to generate representations of larger pieces of text, loosing word order as in bag-of-words approaches, or instead use sequences with the word embeddings themselves as the inputs to be processed by the learning models.

Different machine learning methods have been used in NLP applications, and deep neural networks are nowadays a popular choice. In general, neural networks can be seen as as nested composite functions that transform vector representations, and whose parameters can be trained directly to minimize a given loss function computed over the outputs and the expected results. This training procedure involves an algorithm known as back-propagation, in combination with some variation of gradient descent optimization~\citep{primer}. An optimization procedure that has been frequently used to train deep neural networks is the Adaptive Moment Estimation (Adam) algorithm~\cite{kingma2014adam}. Adam computes parameter updates leveraging an exponentially decaying average of past gradients, together with adaptive learning rates for each parameter. In practice, it performs larger updates for infrequent parameters, and smaller updates for frequent parameters.

The Multi-Layer Perceptron (MLP) is a simple neural network architecture, which consists of a set of nodes forming the input layer, one or more hidden layers of computation nodes, and an output layer of nodes. The input signal propagates through the network layer-by-layer in a feed-forward manner, until it reaches the output node(s). Considering a single hidden layer, the corresponding computations can be written as shown in Equation~\ref{eq:mlp}.
\begin{equation}
\boldsymbol{y}=\mathrm{\sigma} \left( \mathrm{\sigma'}( \boldsymbol{x} \cdot \boldsymbol{A}+ \boldsymbol{a} ) \cdot \boldsymbol{B} + \boldsymbol{b} \right)
\label{eq:mlp}
\end{equation}
In the previous equation, ${\boldsymbol{x}}$ is a vector of inputs and  ${\boldsymbol{y}}$ a vector of outputs. The matrix ${\boldsymbol{A}}$ represents the weights of the first layer and ${\boldsymbol{a}}$ is a bias vector for the first layer, while ${\boldsymbol{B}}$ and ${\boldsymbol{b}}$ are, respectively, the weight matrix and the bias vector of the second layer. The functions $\mathrm{\sigma'(.)}$ and $\mathrm{\sigma(.)}$ both denote an element-wise non-linearity, i.e. the activation functions respectively associated to nodes in the hidden layer, and in the output layer of the network. The softmax function (i.e., a normalized exponential function that produces as output a probability distribution) is often used as the activation in the final layer of a MLP classifier, training the network to minimize a cross-entropy loss defined over the predictions and the ground-truth labels.

While MLPs have been extensively used in NLP applications, the problems in this area often involve capturing regularities over structured data of arbitrary sizes (e.g., sequences of word embeddings). In many cases, this means encoding the structure as a fixed width vector, which we can then be used for further processing. Other network architectures besides MLPs are thus commonly used in NLP, in order to transform a sequence of word embeddings $\boldsymbol{x}_1, \ldots , \boldsymbol{x}_T \in \mathbb{R}^d$ into a vector (e.g., a sentence representation) $\boldsymbol{s} \in \mathbb{R}^h$. These include Convolutional Neural Networks (CNNs) and different forms of Recurrent Neural Networks (RNNs).

CNNs involve the application of $h$ filters, sliding them over the input sequence. Each filter performs a local convolution (i.e., an element-wise matrix multiplication followed by a summation) on the sub-sequences of the input, to obtain a set of feature maps. Then, a global average- or max-pooling over time is performed to obtain a scalar, and the scalars from the $h$ filters are finally concatenated into the sequence representation vector $\boldsymbol{s} \in \mathbb{R}^h$.

Assuming an input sequence $\boldsymbol{x}_1,\ldots,\boldsymbol{x}_T$ (e.g., a sequence of embeddings for a document with $T$ words), a convolution layer of width $k$ works by moving a sliding window of size $k$ over the sequence, creating several instances of windows $\boldsymbol{w}_i=[\boldsymbol{x}_i;\boldsymbol{x}_{i+1};\ldots;\boldsymbol{x}_{i+k-1}]$. A filter, i.e. a linear transformation followed by an activation function, is then applied to each window, resulting in $m$ vectors $\boldsymbol{p}_1$,..., $\boldsymbol{p}_m$ where each is defined as follows:
\begin{equation}
\boldsymbol{p}_i = \mathrm{\sigma}(\boldsymbol{w}_i \cdot \boldsymbol{A} + \boldsymbol{a})
\label{eq:conv}
\end{equation}
In the previous equation, $\mathrm{\sigma(.)}$ is an activation function that is applied element-wise, while \textbf{A} and \textbf{a} are parameters of the network. The $m$ vectors are then passed through a max-pooling layer and a final representation vector \textbf{r} is obtained. Each element $j$ of \textbf{r} is obtained as follows, where $\boldsymbol{p}_i[j]$ denotes the $j$-th component of $\boldsymbol{p}_i$: 
\begin{equation}
\boldsymbol{r}[j] = \max_{1 < i \leq m} \boldsymbol{p}_i[j].
\label{eq:maxpool}
\end{equation}

RNNs can, in turn, be seen as time-dependent neural networks, which at time step $t$ (i.e., at position $t$ for a given input sequence) compute a hidden state vector $\boldsymbol{h}_t$, which is obtained by a non-linear transformation with two inputs, i.e. the previous hidden state $\boldsymbol{h}_{t-1}$ and the current word input $\boldsymbol{x}_t$. The most elementary RNN is often called the Elman RNN, corresponding to the following equation:
\begin{equation}
\boldsymbol{h}_t = \sigma \left( \boldsymbol{W}_1 \cdot \boldsymbol{h}_{t-1} + \boldsymbol{W}_2 \cdot \boldsymbol{x}_t \right)
\end{equation}

However, previous research has noted that the Elman RNN has difficulties in modeling long sequences. Extensions have been proposed to handle this problem, and two well-known examples are Long Short-Term Memory (LSTM) units~\citep{Hochreiter:1997:LSM:1246443.1246450} and Gated Recurrent Units (GRUs)~\cite{chung2014empirical}. Both these approaches involve different components, i.e. gating mechanisms, which interact in a particular way in order to combine previous states with the current inputs. For instance, GRUs correspond to the following equations:

\begin{equation}
\mathbf{\mathbf{z}}_{t} = \mathbf{\varphi_{g}} \left(\mathbf{W}_z \cdot \mathbf{x}_{t} + \mathbf{U}_z \cdot \mathbf{h}_{t-1} + \mathbf{b}_{z} \right)
\end{equation}
\begin{equation}
\mathbf{r}_{t}  = \mathbf{\varphi_{g}} \left( \mathbf{W}_{r} \cdot \mathbf{x}_{t} + \mathbf{U}_{r} \cdot \mathbf{h}_{t-1} + \mathbf{b}_{r} \right)
\end{equation}
\begin{equation}
\mathbf{\tilde{h}}_{t} = \mathbf{\varphi_{h}} \left( \mathbf{W}_{h} \cdot \mathbf{x}_{t} + \mathbf{U}_{h} \cdot ( \mathbf{r}_{t} \odot \mathbf{h}_{t-1})+ \mathbf{b}_{h} \right)
\end{equation}
\begin{equation}
\mathbf{h}_{t} = \mathbf{\mathbf{z}}_{t} \odot \mathbf{h}_{t-1} + (1- \mathbf{\mathbf{z}}_{t}) \odot \mathbf{\tilde{h}}_{t}
\label{eq:gru}
\end{equation}

In the previous equations, the operator $\odot$ denotes the Hadamard product (i.e., the entry-wise product of two matrices), while $\mathbf{x_{t}}$ denotes the input vector at time step $t$, and $\mathbf{h}_{t}$ denotes the hidden state at time step $t$. The activations $\mathbf{\varphi_{g}}(.)$ and $\mathbf{\varphi_{h}}(.)$ are typically chosen to be the sigmoid and the hyperbolic tangent functions, respectively, while the multiple parameters $\mathbf{W}$, $\mathbf{U}$ and $\mathbf{b}$ denote the different weight matrices and bias vectors, adjusted when training the neural network. 

Notice that a GRU involves two gates, namely a reset gate $\mathbf{r}$, that determines how to combine the new input with the previous memory, and an update gate $\mathbf{z}$, that defines how much of the previous memory to keep around. If we set the reset gate to all ones, and the update gate to all zeros, we again arrive at the Elman RNN model that was discussed previously. The gating mechanism allows GRUs to better handle long-term dependencies. By learning the parameters for its gates, the network learns how its internal memory should behave, given that the gates define how much of the input and previous state vectors should be considered.

LSTMs are an alternative type of RNN, with more parameters than GRUs (e.g., they have an extra gate) although also reported to outperform them when more training data is available, and in tasks requiring modeling longer-distance relations~\cite{yin2017comparative}. LSTMs correspond to the following equations, where $\odot$, $\mathbf{\varphi_{g}}(.)$, $\mathbf{\varphi_{h}}(.)$ $\mathbf{x_{t}}$, $\mathbf{h}_{t}$, $\mathbf{W}$, $\mathbf{U}$ and $\mathbf{b}$ share the same meaning as in the GRU equations.

\begin{equation}
\mathbf{\mathbf{i}}_{t} = \mathbf{\varphi_{g}} \left(\mathbf{W}_i \cdot \mathbf{x}_{t} + \mathbf{U}_i \cdot \mathbf{h}_{t-1} + \mathbf{b}_{i} \right)
\end{equation}
\begin{equation}
\mathbf{f}_{t}  = \mathbf{\varphi_{g}} \left( \mathbf{W}_{f} \cdot \mathbf{x}_{t} + \mathbf{U}_{f} \cdot \mathbf{h}_{t-1} + \mathbf{b}_{f} \right)
\end{equation}
\begin{equation}
\mathbf{\mathbf{o}}_{t} = \mathbf{\varphi_{g}} \left(\mathbf{W}_o \cdot \mathbf{x}_{t} + \mathbf{U}_o \cdot \mathbf{h}_{t-1} + \mathbf{b}_{o} \right)
\end{equation}
\begin{equation}
\mathbf{\mathbf{g}}_{t} = \mathbf{\varphi_{h}} \left(\mathbf{W}_g \cdot \mathbf{x}_{t} + \mathbf{U}_g \cdot \mathbf{h}_{t-1} + \mathbf{b}_{g} \right)
\end{equation}
\begin{equation}
\mathbf{c}_{t} = \mathbf{\mathbf{f}}_{t} \odot \mathbf{c}_{t-1} + \mathbf{\mathbf{g}}_{t} \odot \mathbf{i}_{t}
\end{equation}
\begin{equation}
\mathbf{\mathbf{h}}_{t} = \mathbf{\varphi_{h}}(\mathbf{c}_{t}) \odot \mathbf{\mathbf{o}}_{t}
\end{equation}

Notice that LSTMs apply different gating mechanisms, namely through the use of (i) a forget gate that controls how much of the previous memory will be kept, (ii) an input gate that controls how much of the proposed update $\mathbf{\mathbf{g}}_{t}$ should be kept, and (iii) an output gate that controls the output at time $t$.

Both LSTMs and GRUs have been shown to achieve remarkable performance on many sequential learning problems~\cite{yin2017comparative}, particularly when processing short texts. In addition, hierarchical arrangements of LSTMs or GRUs can be used for modeling long documents, accounting not only with word order but also with sentence structure~\cite{hierarchical,franciscoduarte}. For instance Yang et al.~\cite{hierarchical} proposed a hierarchical attention network for document classification, considering documents to be composed of sentences, and modeling sentences as sequences of words. They used a bi-directional GRU (i.e., they concatenate the states produced by two GRUs to generate the result for each time step, with one GRU processing the input in the forward direction, and the other processing the input in reverse) to encode each word in a sentence, followed by an attention mechanism to weight relevant words in the aggregated representation of each sentence. The attention mechanism corresponds to the following equations, effectively weighting the contribution of each word in the sentence, when building the representation:
\begin{equation}
\boldsymbol{u}_{it} = \mathrm{tanh}(\boldsymbol{W}_h \cdot \boldsymbol{h}_{it} + \boldsymbol{b}_w)
\end{equation}
\begin{equation}
\alpha_{it} = \frac{\exp(\boldsymbol{u}_{it}^\top\boldsymbol{u}_w)}{\sum\nolimits_t\exp(\boldsymbol{u}_{it}^\top\boldsymbol{u}_w)}
\end{equation}
\begin{equation}
\boldsymbol{s}_i = \sum_t \alpha_{it} \times \boldsymbol{h}_{it}
\label{eq:hierarchicalattention}
\end{equation}
The vector $\boldsymbol{h}_{it}$ corresponds to the word representation produced by the bi-GRU at time $t$, when processing sentence $i$, while $\boldsymbol{W}_h$ and $\boldsymbol{b}_w$ are parameters to be learned. The vector $\boldsymbol{u}_{it}$ is a hidden representation of $\boldsymbol{h}_{it}$,  $\boldsymbol{u}_w$ is a word-level context vector to be learned, $\alpha_{it}$ is an importance weight, and $\boldsymbol{s}_i$ is the sentence vector, calculated as the weighted sum of the word representations. To create a document vector, the same methodology can be applied. A bi-directional GRU encodes the sentence vectors, and an attention mechanism is used to determine the importance of each sentence, yielding a document vector. The equations for the document-level attention mechanism are analogous to those from Equations 15 to 17. To generate the final prediction, the document vector is processed through feed-forward layer(s) with a final a softmax activation.

Besides document classification, another NLP task that is strongly related to the subject of this article, and where deep learning has been extensively employed, is Natural Language Inference (NLI). In brief, NLI concerns with determining if a given  hypothesis sentence \textit{h} can be inferred from a premise sentence \textit{p}. Generalized versions of the task have also considered multiple possible relations between the hypothesis and the premise (e.g., relations like entailment, contradiction, or neutrality), and large datasets such as the Stanford Natural Language Inference (SNLI)~\cite{snli} or the 
Multi-Genre Natural Language Inference (MultiNLI)~\cite{inference} corpora have enabled significant progress in terms of deep learning methods for building effective semantic representations of natural language information (e.g., models trained for NLI tasks can provide rich domain-general semantic representations).

For instance, Conneau et al.~\cite{supervised} demonstrated that the supervised training of sentence embeddings, based on the aforementioned NLI datasets, can consistently outperform other state-of-the-art approaches for representing sentences in different NLP tasks (e.g., taking the average of \textit{word2vec}~\cite{word2vec} or GloVe~\cite{glove} embeddings, using unsupervised methods such as skip-thought sentence embeddings~\cite{skip}, or using other supervised models such as the \textit{paragram-phrase} approach described by Wieting et al. \cite{paragram}). The model parameters obtained when solving the NLI problem can thus be used for initializing other NLP models. Conneau et al. have also advanced a generic architecture for addressing the NLI task, illustrated in Figure 1. In brief, the premise and the hypothesis can both be encoded by a CNN or an RNN, creating a vector representation for each sentence. These representations are then matched in some way (e.g., through a concatenation of the vectors, through the vector difference, and/or through an element-wise product), fed into a set of fully-connected layers, and finally processed through a feed-forward layer with a softmax activation, that generates a final prediction.

\begin{figure}[t]
  \begin{center}
  \includegraphics[width=0.9\textwidth]{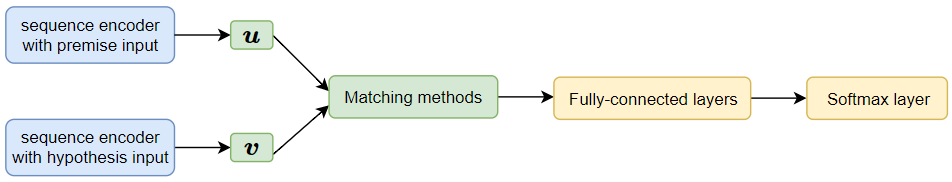}
  \caption{Overview of a generic NLI approach, adapted from the description by Conneau et al. \cite{supervised}.}
  \label{fig:nli}
  \end{center}
\end{figure}

As a sentence encoder (i.e., to generate sentence representations $\boldsymbol{u}$ and $\boldsymbol{v}$, respectively from the premise and the hypothesis), Conneau et al. used a bi-directional LSTM, followed by a max-pooling operation over the sequence of output states. Then, the authors combined both these representations through the concatenation of $\boldsymbol{u}$ and $\boldsymbol{v}$, the element-wise product, and the difference between the vectors (i.e., they used a combination of all three operations). Finally, the result was fed to multiple fully-connected layers, with a final softmax activation. 

Besides the aforementioned general model, several other NLI approaches have been proposed in the literature~\cite{shortcut,gong2017natural,chen2017enhanced}, and evaluated over the SNLI and MultiNLI corpora (e.g., the authors of SNLI maintain a leader-board on the website\footnote{\scriptsize{\url{http://nlp.stanford.edu/projects/snli/}}} describing the corpus). For instance, Nie and Bansal~\cite{shortcut} developed a model that scores an accuracy of 86.1\% on the SNLI test set, an acuracy of 74.6\% on the MultiNLI matched testing set, and 73.6\% on the MultiNLI mismatched testing set. Following the same general approach from Figure 1, these authors used a sentence encoder that takes as input pre-trained GloVe embeddings~\cite{glove} for the words in the sentence, and processes/refines the embeddings through three layers of bi-directional LSTMs with shortcut connections (i.e., the input sequences for the $i$-th bi-LSTM layer are the concatenated outputs of all the previous layers, plus the original word embedding sequence). The final representations of the premise and hypothesis are generated through a max-pooling operation between all the hidden states of the last bi-directional LSTM. A similar approach is used in the present study to encode sentences in the stance detection problem from the Fake News Challenge, although also using inner-attention as in the study by Yang et al. \cite{hierarchical}, instead of just using max-pooling.

\subsection{Stance Detection in the Context of the Fake News Challenge}

The Fake News Challenge (FNC-1) consisted of a competition for evaluating NLP methods designed to solve a stance detection problem between the headline of a news article, and the main body of text for a news article. The leader-board of the competition points to several interesting methods, some of them reviewed in this section. 

The organizers of the competition also provided a baseline consisting of a gradient boosting classifier leveraging hand-crafted features, such as multiple similarity scores between the headline and the body (e.g., based on $n$-gram overlap). Using 10-fold cross validation, the baseline achieved a weighted accuracy of 75.20\%, following the evaluation metric detailed in Section 4.1 (i.e., an accuracy metric that gives extra weight to some of the classes). 

The first place in the competition was obtained by the team \textit{SOLAT in the SWEN}\footnote{\scriptsize{\url{http://blog.talosintelligence.com/2017/06/talos-fake-news-challenge.html}}}, which used an ensemble of two sub-models, each outputting predictions associated with a confidence score. The aggregated model made its decisions with basis on the weighted average of both sub-models, achieving a weighted accuracy of 82.02\%. One of the sub-models was a gradient boosting classifier similar to that of the baseline, leveraging features like the number of overlapping words between the headline and the body, or similarity scores computed from word $n$-gram representations. The second sub-model was based on Convolutional Neural Networks (CNNs) for encoding the headline and the body, using pre-trained \textit{word2vec} embeddings~\cite{word2vec}. The outputs for the headline CNN and body CNN were concatenated and put through feed-forward layers. 

In second place came \textit{Team Athene}\footnote{\scriptsize{\url{http://medium.com/@andre134679/team-athene-on-the-fake-news-challenge-28a5cf5e017b}}}, with a weighted accuracy of 81.97\% and using an ensemble of five identical sub-models with randomly initialized parameters. These sub-models were Multi-layer Perceptrons (MLPs) with seven hidden layers and a softmax activation at the end. Seven distinct feature types were considered, computed from either the headline, the body, or a combination of both. Examples of features include vectors of unigram occurrences, or the cosine distance between headline and body representations computed from a factorization of the unigram occurrence matrix (i.e., from representations produced through non-negative matrix factorization). The final prediction came from hard voting between the five sub-models in the ensemble.

In third place, team \textit{UCL Machine Reading} \cite{riedel2017simple} achieved a weighted accuracy of 81.72\%. This model was also fairly simple, consisting of a MLP with one hidden layer and a final layer with a softmax activation. The input feature vector resulted from a concatenation of Term-Frequency (TF) vectors built from the headline and the body, with the cosine similarity between TF-IDF vectors for the headline and the body.

When considering the top three teams, it is important to notice that all of them leveraged hand-crafted features, together with other neural approaches. In a recent publication, Tosik et al.~\cite{tosikdebunking} reported on tests evaluating the individual contribution of hand-crafted features, using them to feed a single gradient boosting classifier (when directly considering the four class classification problem), or an ensemble of gradient boosting classifiers, where the instances are first classified as either \textit{unrelated} or \textit{related}, and then the \textit{related} instances are assigned to one of the remaining three classes (i.e., \textit{agrees}, \textit{disagrees}, or \textit{discusses}).

Specifically, Tosik et al. leveraged text similarity features such as $n$-gram or word overlap, or the cosine similarity between TF-IDF vectors representing the headline and the body, as well as other miscellaneous features such as the presence of several refuting words in the headline/body, or the length of the headline/body. After achieving a weighted accuracy score of 78.63\% on the FNC-1 testing dataset, the authors conducted ablation tests and concluded that the most helpful features were the overlap features between $n$-grams and words of the headline and body, the refuting features based on a lexicon, and distance measures such as the cosine similarity between TF-IDF vectors of the headline and the body, and the Word Mover's Distance~\cite{wmd} between the headline and body text. On the other hand, sentiment features and other distance scores, such as the Hamming distance, did not contribute to the obtained result.

Pfohl and Legros~\cite{stanfordattention} reported on other interesting submissions to FNC-1. These authors experimented with four different approaches, namely (i) a bag-of-words method, (ii) a basic LSTM approach, (iii) a method leveraging an LSTM together with an inner-attention mechanism, and (iv) a more sophisticated approach which the authors named conditional encoding LSTM with attention (CEA-LSTM).  

The bag-of-words model was built by averaging embeddings for words occurring in the headline, for words occurring in the body, and then concatenating these vectors. This result is processed by a feed-forward network with a softmax output layer. Both the basic LSTM model, and the model combining an LSTM with an inner-attention mechanism, processed a concatenation of the headline and the article body to classify the stance. The inner-attention was computed over a window with the first 15 tokens of the concatenated text. Finally, the CEA-LSTM processed the headline and the body with two separate LSTMs, using the final hidden state of the headline LSTM as the first hidden state of the body LSTM. An attention mechanism operated over a  window with the last 15 output states from the headline LSTM, together with the final hidden state of the body LSTM. 
The best results reported by Pfohl and Legros were achieved with the CEA-LSTM method, corresponding to a weighted accuracy of 80.8\% (i.e., the method based on LSTMs failed to outperform other simpler approaches).

More recently, after the announcement of the winners for FNC-1, Bhatt et al.~\cite{combination} described an approach based on a MLP for combining neural representations, statistical summaries of the data, and feature engineering heuristics. The neural representations leveraged skip-thought vectors~\cite{skip} (i.e., sentence representations learned in an unsupervised manner, by encoding a sentence to predict the sentences around it in a given text corpora) to encode the headline and the body. Given the encodings for the headline and the body, the component-wise product and the absolute difference between the vectors were computed and used as features. The statistical features correspond to unigram occurrences within the headline and the body, weighted according to TD-IDF. Finally, the external heuristic features included the number of similar words in the headline and the body, the cosine similarity between vector encodings for the  headline and the body, the number of matching $n$-grams, the sentiment difference between the headline and the body, etc. Each of the three main sets of features is processed through feed-forward layers before being combined and processed by a final layer with a softmax activation. This approach achieved a weighted accuracy of 83.08\%, scoring higher than the winning team in FNC-1. To the best of our knowledge, this result corresponds to the current state-of-the-art in the FNC-1 task.

Chaudhry et al. \cite{stanfordstance} described six different models for stance detection, which were also tested on FNC-1 data. The first model was a baseline leveraging the Jaccard similarity between unigrams from the headline and from the sentences in the body. The second model was a MLP leveraging representations for the headline and the body, e.g. obtained by averaging pre-trained GloVe embeddings~\cite{glove}. The third model explored the use of two independent LSTMs, one encoding the headline and another encoding the body. The two final state vectors are passed to a softmax layer that generates the final prediction. The fourth model also used an LSTM to encode the headline, but now its final state vector was used to initialize another LSTM that encodes the body. The fifth approach augmented the fourth model by considering bi-directional LSTMs. Finally, the sixth approach extended the fifth model with a self-attention mechanism on top of the encoder LSTMs. All six models were tested on a custom training/testing data split (i.e., the official test set was not available at the time the paper was written, and thus the results cannot be directly compared against those reported for the FNC-1 competition). The best results were achieved by the fifth model (i.e., the bi-directional encoder), corresponding to a weighted accuracy of 95.3\%.

Similarly to Tosik et al.~\cite{tosikdebunking}, 
Bourgonje et al. \cite{clickbait} also addressed the separate tasks of (a) determining whether a headline-body pair is \textit{unrelated} or \textit{related} and, in the latter case, (b) the task of determining whether the body \textit{agrees}, \textit{disagrees}, or \textit{discusses} the subject of the headline. Again, experiments leveraged data from FNC-1. In the first task, the authors start by gathering two sets of $n$-grams, one for the headline and another for the body. Then, the number of matching $n$-grams is multiplied by the length and IDF value of the matching $n$-grams, and divided by the total number of $n$-grams. If the resulting score is above a threshold, the pair is considered to be \textit{related}. For the second task (i.e., the three class classification problem), Bourgonje et al. used a logistic regression classifier trained on features extracted from the headlines of the FNC-1 dataset. If the distance between the best and the second best scoring classes is below a given threshold, the authors use a separate binary classifier, trained on features from both the headlines and the bodies of FNC-1 news articles. Three binary classifiers were trained for this second-level model, namely one for discriminating between \textit{agrees} and \textit{disagrees}, one for \textit{agrees} versus \textit{discusses}, and another one for \textit{discusses} versus \textit{disagrees}. Bourgonje et al. evaluated the proposed approach by leveraging 50 different tests with random 90-10 splits of the FNC-1 dataset, scoring a weighted accuracy of 89.59\%. However, these results are again not directly comparable to those from the FNC-1 participants.

Zeng et al. \cite{stanfordneural} tested six different encoders in a neural method for the FNC-1 task, based on concatenating representations for the headline and the body, afterwards generating the final classification through a softmax layer. The first encoder consisted of separate bi-directional GRUs, one for the headline and another for the body, with representations obtained from the final GRU states. In the second approach, the headline is fed to a bi-directional GRU, whose final hidden state is used to initialize another bi-directional GRU that processes the body. The third encoder uses a bi-directional GRU to process the concatenation of the headline and the body text.

The fourth and fifth models both leverage attention mechanisms. Specifically, the fourth model compares every state of the bi-directional GRU processing the body with the final hidden state of the bi-directional GRU that processes the headline. The attention weights are generated as follows:

\begin{equation}
\begin{split}
\alpha_i = \mathrm{softmax}(\boldsymbol{q}^\top \cdot \boldsymbol{W}_s \cdot \widetilde{\boldsymbol{p}}_i).
\label{eq:attentivereader}
\end{split}
\end{equation}

In the previous equation, $\boldsymbol{q}$ is the final state of the headline bi-directional GRU, $\widetilde{\boldsymbol{p}}_i$ is the $i$-th hidden state of the body bi-directional GRU, $\boldsymbol{W}_s$ are trainable weights, and $\alpha_i$ is the attention weight of the $i$-th body token. The attention weights are used to compute a weighted sum of the hidden states from the body bi-directional GRU, and this result is then concatenated with the final hidden state of the bi-directional GRU encoding the headline.

The fifth model is very similar to the fourth, in this case computing the attention weights by comparing the hidden states of the body GRU with all the hidden states of the headline GRU, instead of only the final state. The final attention weight for each body hidden state is given by the maximum value over all the weights calculated for every hidden state of the headline GRU.

Finally, the sixth approach was named the Bilateral Multiple Perspective Matching model. First, the headline and the body are encoded using separate bi-directional GRUs. Then, a cosine similarity is calculated between every hidden state from the headline GRU, and every hidden state from the body GRU:
\begin{equation}
\begin{split}
\boldsymbol{m}_k = \mathrm{cosine}(\boldsymbol{w}_k \odot \boldsymbol{v}_1, \boldsymbol{w}_k \odot \boldsymbol{v}_2).
\label{eq:cosinesim}
\end{split}
\end{equation}
In the previous equation, $\boldsymbol{w}_k$ are trainable weights, and $\boldsymbol{v}_1$ and $\boldsymbol{v}_2$ represent the hidden states to be compared. For every headline hidden state, several representations were created (i.e., as many $\boldsymbol{m}_k$ vectors as the length of the body). A max-pooling is then applied, in order to obtain a single vector for every headline state. For the body hidden states, the process is analogous. After this first layer of representations, a second layer again uses bi-directional GRUs to further process the headline and the body representations. The two resulting hidden states are finally concatenated, and fed to feed-forward layers that perform the classification.

The aforementioned six models were evaluated on a custom split of the FNC-1 dataset, and the fifth approach described by the authors (i.e., the attention model comparing the body hidden states with every headline hidden state) achieved the best results, with a weighted accuracy of 85.2\%. Again, this result cannot be directly compared against those from the FNC-1 participants.

More recently, Mohtarami et al. \cite{Mohtarami2018} introduced a memory network for stance detection, evaluating it on the FNC-1 testing dataset. The proposed memory network can be seen as 6-tuple $\{M, I, F, G, O, R\}$, where the memory $M$ is a sequence of representations, $I$ is a mapping from inputs to their representations, $F$ is an inference component that identifies the relevant parts of the input, $G$ is a generalization component that updates the memory according to $F$, $O$ is an output generated for each new input given the current memory state, and finally  $R$ is conversion from $O$ into a desired response format.

Given a news article and a headline, the input component $I$ first converts the news article into a 3D tensor $\boldsymbol{d} = (\boldsymbol{X}, \boldsymbol{W}, \boldsymbol{E})$, where $\boldsymbol{X} = \{\boldsymbol{x}_1, \dots, \boldsymbol{x}_n\}$ is a set of paragraphs that constitute the document, $\boldsymbol{W} = \{\boldsymbol{w}_1, \dots, \boldsymbol{w}_v\}$ are the words that represent each paragraph $\boldsymbol{x}_j$, and $\boldsymbol{E} = \{\boldsymbol{e}_1, \dots, \boldsymbol{e}_v\}$ are the word embeddings. Every $\boldsymbol{x}_j$ is separately processed through an LSTM and a CNN, generating representations $\boldsymbol{m}_j$ and $\boldsymbol{c}_j$. A similar procedure is considered for the headline, generating representations $\boldsymbol{s}_{lstm}$ and $\boldsymbol{s}_{cnn}$.

The inference component $F$ takes the previously computed representations and generates two similarities:

\begin{equation}
\boldsymbol{p}^j_{lstm} = \boldsymbol{s}_{lstm}^\top \cdot \boldsymbol{M} \cdot \boldsymbol{m}_j
\end{equation}
\begin{equation}
\boldsymbol{p}^j_{cnn} = \boldsymbol{s}_{cnn}^\top \cdot \boldsymbol{M'} \cdot \boldsymbol{c}_j\\
\label{eq:sims}
\end{equation}
In the previous equations, $\boldsymbol{M}$ and $\boldsymbol{M'}$ are trainable similarity matrices. Additionally, another similarity vector $\boldsymbol{p}^j_{tfidf}$ is computed by applying the cosine similarity metric to TF-IDF representations for the news headline and for each sentence of the news article body.

The memory component $M$ and the generalization component $G$ update the $\boldsymbol{m}_j$ vectors in the following way:
\begin{equation}
\begin{split}
\boldsymbol{\tilde{m}}_j = \boldsymbol{m}_j \odot \boldsymbol{p}^j_{tfidf}
\end{split}
\end{equation}
Then, the updated $\boldsymbol{\tilde{m}}_j$ and $\boldsymbol{s}_{lstm}$ are used by the inference component $F$ to compute separate $\boldsymbol{\tilde{p}}^j_{lstm}$ vectors, using the same procedure detailed above. The new $\boldsymbol{\tilde{p}}^j_{lstm}$ vectors are then used to update the $\boldsymbol{c}_j$ vectors:
\begin{equation}
\begin{split}
\boldsymbol{\tilde{c}}_j = \boldsymbol{c}_j \odot \boldsymbol{\tilde{p}}^j_{lstm}
\end{split}
\end{equation}
Finally, the updated $\boldsymbol{\tilde{c}}_j$ representations in conjunction with $\boldsymbol{s}_{cnn}$ are leveraged to compute new $\boldsymbol{\tilde{p}}^j_{cnn}$ vectors, using the same procedure as explained above. All the aforementioned vectors are used in the output component $O$ to compute the following vector:

\begin{equation}
\small
\boldsymbol{o} = \mathlarger{\mathlarger{[}}\mathrm{mean}(\{\boldsymbol{c}_j\}); 
\mathrm{max}(\{\boldsymbol{p}^j_{cnn}\}); \mathrm{mean}(\{\boldsymbol{p}^j_{cnn}\}); 
\mathrm{max}(\{\boldsymbol{p}^j_{lstm}\}); \mathrm{mean}(\{\boldsymbol{p}^j_{lstm}\}); \mathrm{max}(\{\boldsymbol{p}^j_{tfidf}\}); \mathrm{mean}(\{\boldsymbol{p}^j_{tfidf}\}) \mathlarger{\mathlarger{]}}
\end{equation}

In the response component $R$, the concatenation $[\boldsymbol{o}; \boldsymbol{s}_{lstm}; \boldsymbol{s}_{cnn}]$ is fed to feed-forward layers, and a final softmax layer gives the prediction for the stance. In their tests, Mohtarami et al. achieved a weighted accuracy of 81.23\%, again attesting to the difficulties associated to the application of modern neural approaches to the FNC-1 task.

\section{Proposed Neural Network Architecture}

Our neural network architecture takes inspiration on the models by Yang et al.~\cite{hierarchical} and by Nie and Bansal~\cite{shortcut}, described on the previous section. Figure 2 provides an high-level overview of the proposed approach. 

The headline is processed through the sentence encoder described in Section 3.1, which outputs the corresponding representation. In turn, the body of the news article is processed through the document encoder described in Section 3.2, which leverages a two-layer hierarchical model combining the sentence encoder with an encoder for the sequence of sentences. A third optional branch compares the headline with the first two sentences of the body, leveraging the sentence encoder to build the involved representations. This matcher, as well as the common sentence encoder, can be pre-trained with the SNLI and MultiNLI datasets, as explained in Section 3.3.

The representations for the headline, the entire body, and the first two sentences from the body, are matched through vector operations such as the element-wise product, the vector difference, or the concatenation. These results are then combined with external features computed from the headline and/or the body -- see Section 3.4 -- and the result is processed by two fully-connected layers, with the final one producing the stance classification. The following sections detail the components of this architecture.

\begin{figure}[t]
  \begin{center}
  \includegraphics[width=1\textwidth]{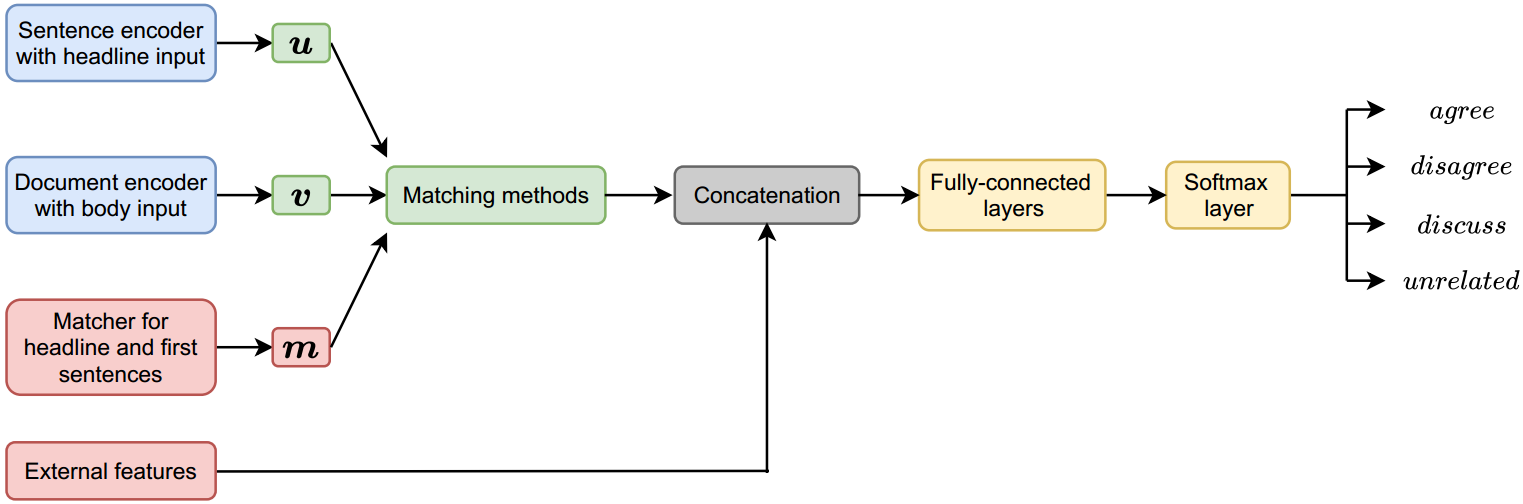}
  \caption{Overview of the proposed approach.}
  \label{fig:baseline}
  \end{center}
\end{figure}

\subsection{Encoding Sentences with Bi-Directional RNNs and Summarization Mechanisms}

The proposed neural architecture takes inspiration from Conneau et al.~\cite{supervised}, leveraging a sentence encoder responsible for building representations for input sentences, e.g., for the news headlines. This encoder takes as input a sequence of $w_l$ words (i.e., a sentence), with $l \in [0, L]$ and where $L$ is the length of the sequence, replacing each word by a pre-trained GloVe embedding~\cite{glove}. The resulting matrix of embedded words $\boldsymbol{E}$ is processed through a bi-directional RNN, which  generates a hidden state matrix $\boldsymbol{H}_1$ as follows:
\begin{equation}
\boldsymbol{h}_{tf} = \mathrm{RNN_{forward}}(\boldsymbol{E}_t), t \in [1, \ldots, L] 
\end{equation}
\begin{equation}
\boldsymbol{h}_{tb} = \mathrm{RNN_{backward}}(\boldsymbol{E}_t), t \in [1, \ldots, L]
\end{equation}
\begin{equation}
\boldsymbol{h}_t = [\boldsymbol{h}_{tf}; \boldsymbol{h}_{tb}], t \in [1, \ldots, L]
\end{equation}
\begin{equation}
\boldsymbol{H} = [\boldsymbol{h}_1, \ldots, \boldsymbol{h}_L].
\label{eq:bigru1}
\end{equation}

In the previous equations, the RNN function can be instantiated either with the LSTM or with the GRU equations, both shown in Section 2.1 and hence not repeated here. Finally, the RNN states $\boldsymbol{h}_{t}$ from matrix $\boldsymbol{H}$ are processed through a summarization mechanism, which outputs a single vector. For this work, we consider the summarization mechanism to be either a max-pooling operation, and/or an inner-attention mechanism defined as shown in the next equations:
\begin{equation}
\boldsymbol{u}_{t} = \mathrm{tanh}(\boldsymbol{W} \cdot \boldsymbol{h}_{t}),           t \in [1, \ldots, L]
\end{equation}
\begin{equation}
\alpha_{t} = \frac{\exp(\boldsymbol{u}_{t}^\top)}{\sum\nolimits_t\exp(\boldsymbol{u}_{t}^\top)}, t \in [1, \ldots, L]
\end{equation}
\begin{equation}
\boldsymbol{s} = \sum_t \alpha_{t} \times \boldsymbol{h}_{t},  t \in [1, \ldots, L]
\label{eq:attention}
\end{equation}
In the equations above, the matrix $\boldsymbol{W}$ corresponds to trainable weights, $\alpha_t$ is an importance weight assigned to each hidden state, and $\boldsymbol{s}$ is the final representation of the input sentence.

In order to further augment the sentence encoder described above, we also experimented with stacking two layers of bi-directional RNNs, feeding the second bi-directional RNN layer with the concatenation of original embeddings $\boldsymbol{E}$ and the hidden states $\boldsymbol{H}$ from the first bi-RNN (i.e., considering shortcut connections between the two layers). In this case, a second hidden state matrix $\boldsymbol{H}_2$ is produced through the same procedure described through Equations 25-28. Finally, the RNN states $\boldsymbol{h}_{t}$ from matrix $\boldsymbol{H}_2$ are processed through a summarization mechanism, similar to what was detailed in the previous paragraph. 

Figure 3 illustrates the sentence encoder combining all the aforementioned mechanisms, thus using two stacked layers of bi-RNNs (i.e., GRUs or LSTMs), shortcut connections, and a combination of max-pooling and neural attention obtained through a concatenation operation.

\begin{figure}[t!]
  \begin{center}
  \includegraphics[width=1\textwidth]{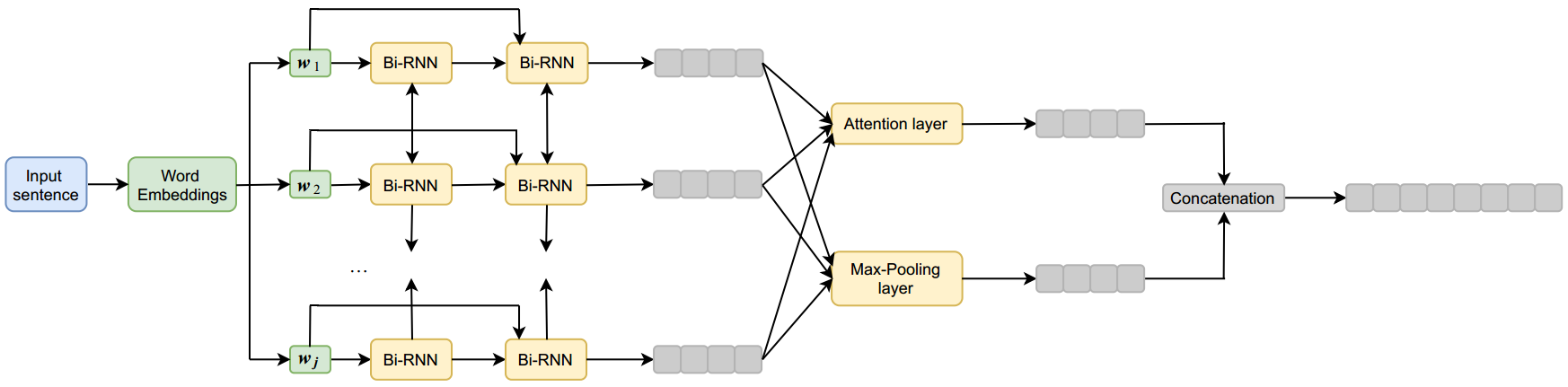}
  \caption{An extension of the previously detailed sentence encoder.}
  \label{fig:headencoder}
  \end{center}
\end{figure}

\subsection{Hierarchical Approach for Encoding Documents}

The neural encoder for the news article body leverages the sentence encoder from the previous section, combining it with the hierarchical approach described by Yang et al.~\cite{hierarchical}. Each sentence in the news article is thus encoded through the procedure described in Section 3.1, and the resulting sequence of sentence vectors is then processed through a similar encoder (i.e., using a bi-directional RNN or a stacked model with two bi-RNNs and shortcut connections, followed by max-pooling and/or an inner-attention mechanism).

\subsection{Model Pre-Training with the SNLI and MultiNLI Datasets}

Besides matching the headline against the entire body of the news article, we also experimented with a neural architecture that additionally attempts to match the headline against the first two sentences of the body, through a separate branch. This approach leverages the intuition that the opening sentences of news articles often contain a summary for the entire document. The sentence encoder described in Section 3.1 is used to represent the headline and the first sentences of news article, and these representations are then combined though vector operations (i.e., the element-wise multiplication, the difference, and the concatenation).

Moreover, instead of randomly initializing all the parameters of the proposed neural network architecture, we used the SNLI~\cite{snli} and MultiNLI~\cite{inference} datasets to pre-train the parts of the network that do (i) the encoding of sentences, and (ii) the matching of the headline against the first two sentences of the body, as described in the previous paragraph. Previous studies have already attested to the benefits of similar pre-training procedures in other natural language processing and text classification tasks~\cite{supervised}.

The headline is seen as the hypothesis sentence from the NLI task, and the first two sentences from the body are seen as the premise. Our complete NLI model thus involves the same  sentence encoder and the aforementioned matching strategy (i.e., combining the representations for the hypothesis and the premise through the element-wise multiplication, vector difference, and vector concatenation), combined with a final softmax layer that returns the NLI class. After pre-training this model, we simply ignore the final softmax layer and re-use the remaining components on our complete FNC-1 neural architecture.

\subsection{Combining the Matched Representations with External Features}

The previous section described the vector operations used for matching the headline against the first two sentences of the body. A similar approach is also used for matching the representation for the  headline against the representation for the entire body of the news article. Thus, the vector resulting from the pre-trained NLI network, which matches the headline against the first two sentences, is concatenated with the vector encoding the entire body, and with the results from the element-wise product and the difference between the vectors representing the headline and the body.

Moreover, taking inspiration on the baseline model proposed by the organizers of FNC-1, we also combine the aforementioned representations built through the neural network with external features computed from the headline and/or the body. The following list enumerates the considered features:

\begin{enumerate}
\item The number of words in common between (a) the headline and the body of the news article, and (b) the headline and the first two sentences of the body;

\item Refutation features, based on the presence of refuting words, listed in a given dictionary, in the headline (e.g., words like {\it deny}, {\it doubt}, {\it fraud} or {\it debunk});

\item Polarity features, based on the presence of words with high emotion/sentiment polarity in (a) the headline, (b) the entire body, and (c) the first two sentences;

\item The number of word tokens that are common to (a) the headline and the body of the news article, and (b) the headline and the first two sentences of the body;

\item The number of word $n$-grams that are common to (a) the headline and the body of the news article, and (b) the headline and the first two sentences of the body;

\item The soft cosine similarity metric~\cite{charlet2017simbow}, computed between representations leveraging word occurrences for (a) the headline and the entire body, or (b) the headline and the first two sentences of the body;

\item The BLEU score~\cite{bleu} computed between (a) the headline and the set of sentences from the body, and (b) the headline and the first two sentences of the body;

\item ROUGE scores~\cite{rouge} computed between (a) the headline and the set of sentences from the body, and (b) the headline and the first two sentences of the body;

\item The CIDEr similarity score~\cite{cider}, computed between (a) the headline and the set of sentences of the article, and (b) the headline and the first two sentences of the body;

\item The cosine similarity metric, computed between TF-IDF vector representations for the words occurring in the headline, and in the body of the news article.

\item A vector representation of the headline, with 50 dimensions, resulting from a Singular Value Decomposition (SVD) of a matrix with TF-IDF representations for the texts;

\item A vector representation for the body of the article, with 50 dimensions, resulting from a Singular Value Decomposition (SVD) of a matrix with TF-IDF representations for the texts;

\item The cosine similarity metric, computed between the SVD vectors for the headline and the body;

\item A vector representation for the headline, with 300 dimensions, produced by averaging the \textit{word2vec} embeddings for the words occurring in the headline;

\item A vector representation for the body, with 300 dimensions, produced by averaging the \textit{word2vec} embeddings for all the words occurring in the body of the article;

\item The cosine similarity metric, computed between averaged \textit{word2vec} embeddings for the headline and body;

\item Sentiment polarity scores for the headline and the body, computed with basis on a word polarity lexicon.
\end{enumerate}

The first 5 features from the previous enumeration were taken from the official FNC-1 baseline system, provided by the organizers. Features 10 to 17 were taken from the system\footnote{\scriptsize{\url{http://github.com/Cisco-Talos/fnc-1}}} that won the FNC-1 challenge, developed by the team \textit{SOLAT in the SWEN}. Finally, in additon to external features based on previous work within the context of FNC-1, we also considered features from previous work concerned with assessing text similarity (e.g., the soft cosine similarity metric), or concerned with the evaluation of NLP methods for automated translation (i.e., the BLEU score), text summarization (i.e., ROUGE scores), or caption generation (i.e., the CIDEr score). The three aforementioned metrics all attempt to assess the similarity between a given input text and a set of reference texts. When computing them, we considered the input text (i.e., the candidate) to be the headline, and the set of reference texts to be formed by each sentence that composes the corresponding news article. When considering only the first two sentences from an article, we instead calculated the scores between the headline (i.e., the candidate) and the concatenation of the referred first two sentences (i.e., a reference set with a single instance).

BLEU~\cite{bleu} works by counting the number of matching unigrams between the candidate text and the references. For every unigram in the candidate, BLEU saves the maximum amount of times it appears in a reference. BLEU then takes the minimum values from the aforementioned counts, and the number of times that the corresponding unigram appears in the candidate, hence generating a value $m$ for every unigram in the candidate. Finally, the BLEU score between the candidate and the references is computed by summing all the $m$ values for every unigram, and then dividing the result by the number of unigrams in the candidate text.

ROUGE~\cite{rouge} is similarly based on matches between the candidate and the references. Different variations of the metric can be computed, and we specifically used ROUGE-1, ROUGE-2, and ROUGE-L. ROUGE-1 is based on the average of the number of overlapping unigrams between the candidate and the references. ROUGE-2 is similar, considering instead the number of overlapping bi-grams. ROUGE-L averages the number of longest common sub-sequences between the candidate and the references.

CIDEr~\cite{cider} begins by representing each sentence (i.e., the candidate and the references) as a TF-IDF vector, with basis on the $n$-grams that compose it. The CIDEr score between a candidate sentence and a set of reference sentences is then computed as follows:

\begin{equation}
\mathrm{CIDEr}(c, S) = \sum\limits_{n=1}^N w_n \times \left( \frac{1}{m} \times \sum\limits_{j} \frac{\boldsymbol{g}^n_c \cdot \boldsymbol{g}^n_{s_j}}{||\boldsymbol{g}^n_c)|| \hspace{0.1cm} \times ||\boldsymbol{g}^n_{s_j}||} \right)
\label{eq:cider}
\end{equation}

In the previous equation, $c$ is the candidate sentence, $S$ is the set of references,  $\boldsymbol{g}^n_c$ is the TF-IDF vector representation for the $n$-grams occurring in the candidate, and $\boldsymbol{g}^n_{s_j}$ is the TF-IDF vector for the $j$-th reference. The number of references is represented by $m$, while $w_n$ is a weight, and $N$ is the maximum length of $n$-grams to consider. In our case, we defined $w_n = 1/N$ and $N = 4$.

Finally, the Soft Cosine Similarity (SCS) metric computes the similarity between two texts leveraging bag-of-words representations, being computed as follows:

\begin{equation}
\begin{split}
\mathrm{SCS}(\boldsymbol{a}, \boldsymbol{b}) = \frac{\boldsymbol{a}^\top \cdot \boldsymbol{M} \cdot \boldsymbol{b}}{\sqrt{\boldsymbol{a}^\top \cdot \boldsymbol{M} \cdot \boldsymbol{a}} \times \sqrt{\boldsymbol{b}^\top \cdot \boldsymbol{M} \cdot \boldsymbol{b}}}
\end{split}
\label{eq:scs}
\end{equation}

In the above equation, $\boldsymbol{a}$ and $\boldsymbol{b}$ are the representations for the input texts, and $\boldsymbol{M}$ is a relation matrix whose element $m_{ij}$ expresses some relation between word $i$ and word $j$, hence guaranteeing that two texts without any word in common have a score above 0 as soon as they share related words. For our work, $\boldsymbol{M}$ is a sparse term similarity matrix computed from 50-dimensional GloVe word embeddings~\cite{glove}.

\section{Experimental Evaluation}

We evaluated the stance detection model described on the previous section by leveraging the problem definition and the evaluation methodology (i.e., the dataset and the metrics) from the Fake News Challenge (FNC-1). This challenge aimed to evaluate models for estimating the stance of a news article body towards a given news headline. One such model should take as input a headline and a body text, from the same article or from different articles, and return as output a classification for the stance, considering the following four categories:
\begin{enumerate}
\item \textbf{Agrees}: The body agrees with the headline;
\item \textbf{Disagrees}: The body disagrees with the headline;
\item \textbf{Discusses}: The body is related with the headline, but it does not take a position regarding its subject;
\item \textbf{Unrelated}: The body discusses a different topic from that of the headline.
\end{enumerate}

\subsection{The FNC-1 Dataset and Experimental Methodology}

To support the training and testing of models participating in the FNC-1, the organizers of the challenge released training and testing datasets. The training dataset contains 49,972 instances (i.e., pairs of headline and body texts) classified with a stance. The labeled testing dataset was, in turn, released after the end of the competition, and it contains 25,419 instances. Table 1 presents elementary characterization statistics for the training/testing datasets provided by the FNC-1 organizers.

Due to the imbalance in the class distribution, and also due to the fact that the distinction between \textit{agree}, \textit{disagree} and \textit{discusses} is much more relevant to fake news detection, the FNC-1 organizers suggested a weighted scoring system. If a test instance is \textit{unrelated} and the model labels it correctly, the score will be incremented by 0.25. If the test instance is related, i.e., the correct label is either \textit{agree}, \textit{disagree} or \textit{discusses}, then the score will be incremented by 0.25 if the model labels the pair with one of the previously mentioned labels. In case the model chooses the correct label of a related test instance, the score will be incremented by an additional 0.75. Summing up, the equation for the proposed weighted accuracy metric is as follows:
\begin{equation}
\begin{split}
Acc_{FNC} = 0.25\times Acc_{Related, Unrelated} + 0.75\times Acc_{Agree, Disagree, Discuss}
\label{eq:eval}
\end{split}
\end{equation}

\begin{table}[t]
  \begin{center}
    \caption{Characterization statistics for the FNC-1 dataset.}
    \label{tab:fnctraining}
    \begin{tabular*}{\textwidth}{l @{\extracolsep{\fill}} c @{\extracolsep{\fill}} c}
    
      \textbf{Property} &  \textbf{Training Split} &  \textbf{Testing Split}\\
      \hline
       Number of instances & 49,972 & 25,413\\
       Number of different news headlines & 1,648 & 893\\
       Number of different news article bodies & 1,683 & 899\\
       \hline
       Headline average length (tokens) & 13 & 12\\
       Body average length (tokens) & 428 & 396\\
       \hline
       Percentage of \textit{unrelated} pairs & 73.131\% & 72.203\%\\
       Percentage of \textit{discuss} pairs & 17.828\% & 17.566\% \\
       Percentage of \textit{agree} pairs & 7.360\% & 7.488\%\\
       Percentage of \textit{disagree} pairs & 1.681\% & 2.742\%\\
       \hline
    \end{tabular*}
  \end{center}
\end{table}

\subsection{Parameters Involved in the Proposed Approach}

We evaluated the complete model outlined in Section 3 against previous alternatives for the FNC-1 task, specifically focusing on previous studies that used the same training/testing splits. We also performed an ablation study, removing some of the components from the full model and seeing how that affects the performance.

Our deep neural network relied on representations for the word tokens based on pre-trained GloVe embeddings with 300 dimensions~\cite{glove}. Out-of-vocabulary words in the testing dataset were represented by the GloVe embedding of their most similar word, as given by the Jaro-Winkler similarity metric~\citep{winkler1990string}.

The model was implemented with the keras\footnote{\scriptsize{\url{http://keras.io/}}} deep learning framework, and the corresponding source code is available on GitHub\footnote{\scriptsize{\url{http://github.com/LuisPB7/fnc-msc}}}. Given the implementations for RNNs available within keras, every sentence (i.e., the headlines, the premise and hypothesis sentences from the NLI datasets, and the sentences within the body of news articles) was either zero-padded or truncated to have 50 tokens. Every news article body was also zero-padded or truncated to have to 30 sentences. We used GRUs and LSTMs with hidden states of 300 dimensions (i.e., the bi-directional RNN representations have 600 dimensions), and the feed-forward layers before the final softmax layer are composed of 600 and 300 neurons, respectively. 

We trained the neural models leveraging the Adam~\cite{kingma2014adam} optimizer with default parameters, for a maximum of 100 epochs. An early stopping was activated in case the training loss does not decrease for 2 consecutive epochs. 

\subsection{The Obtained Results}

\begin{table}[t!]
  \begin{center}
    \caption{Accuracy on the natural language inference task, over the SNLI and MultiNLI testing datasets.}
    \begin{tabular*}{\textwidth}{ l @{\extracolsep{\fill}} c @{\extracolsep{\fill}} c @{\extracolsep{\fill}} c @{\extracolsep{\fill}}}
    
       &  & \multicolumn{2}{c}{\textbf{MultiNLI}}\\
    \cline{3-4}
       \textbf{Method} & \textbf{SNLI} & \textbf{Matched} & \textbf{Mismatched} \\
      \hline
       Most Frequent Class & 34.3 & 36.5 & 35.6\\
       Continuous BOW (Averaging Word Embeddings) & 75.3 & 65.2 & 64.6\\
       BiLSTM & 83.3 & 67.5 & 67.1\\
        Enhanced Sequential Inference Model \cite{chen2017enhanced} & \textbf{88.0} & 72.4 & 71.9\\
       Nie and Bansal \cite{shortcut} & 86.1 & 74.6 & 73.6\\
       Chen et al. \cite{gated} & 85.5 & 74.9 & 74.9\\
       Conneau et al. \cite{supervised} & 85.0 & -- & --\\
       Densely Interactive Inference Network \cite{gong2017natural} & \textbf{88.0} & \textbf{79.2} & \textbf{79.1} \\
       Directional Self-Attention Encoders \cite{disan} & 85.6 & 71.0 & 71.4\\
       Compare-Propagate Alignment-Factorized Encoders \cite{cafe} & 85.9 & 78.7 & 77.9\\
       Gumbel TreeLSTM Encoders \cite{gumbel} & 86.0 & -- & --\\
       Reinforced Self-Attention Network \cite{shen2018reinforced} & 86.3 & -- & --\\
       Distance-Based Self-Attention Network \cite{im2017distance} & 86.3 & 74.1 & 72.9\\
       \hline
       Bi-GRU sentence encoder + attention & 83.2 & 69.4 & 69.5\\
       Bi-LSTM sentence encoder + attention & 83.5 & 69.9 & 70.4\\
       Bi-GRU sentence encoder + max-pooling & 81.6 & 66.5 & 67.3\\
       Bi-LSTM sentence encoder + max-pooling & 84.5 & 70.7 & 71.1\\
       Bi-GRU sentence encoder + max-pooling + attention & 82.8 & 67.2 & 68.2\\
       Bi-LSTM sentence encoder + max-pooling + attention & 84.1 & 70.8 & 70.8\\
       Stacked Bi-GRUs + shortcut connections + max-pooling & 84.0 & 68.5 & 68.9\\
       Stacked Bi-LSTMs + shortcut connections + max-pooling & 84.8 & 71.4 & 72.2\\
       Stacked Bi-GRUs + shortcut connections + max-pooling + attention & 83.2 & 68.9 & 68.4\\
       Stacked Bi-LSTMs + shortcut connections + max-pooling + attention & 84.4 & 70.7 & 70.5\\
       \hline
    \end{tabular*}
  \end{center}
\end{table}

The complete model described in Section 3 involved pre-training the branch that matches the headline against the first two sentences of the body (and also the common sentence encoder used in the remaining branches) with data from the SNLI and MultiNLI corpora. We therefore evaluated the performance of this sub-component against previous alternatives for Natural Language Inference (NLI). The results are given in Table 2, presenting accuracy scores over the SNLI testing dataset, and over the \textit{matched} (i.e., including news subjects that also appear in the training dataset) and \textit{mismatched} portions of the MultiNLI testing dataset.

Besides the complete sentence encoder described in Section 3.1, considering a stacked arrangement of bi-GRUs or bi-LSTMs with shortcut connections, we also evaluated variants leveraging (a) a single bi-directional RNN layer, or (b) using only max-pooling or neural attention as the summarization method, instead of concatenating the results of max-pooling and attention. The best results on the SNLI dataset were obtained with a single bi-directional LSTM combined with max-pooling, whereas for the MultiNLI dataset the best results were achieved with a single bi-LSTM together with a combination of max-pooling and neural attention. Bi-directional GRUs achieved consistently worse results, and the combination of max-pooling with neural attention achieved almost similar results to the usage of max-pooling alone (i.e., better results when leveraging GRUs, and in some cases neural attention lead to worse results when leveraging LSTMs).

Although we outperform simpler NLI baselines, our results are still far from the current state-of-the-art. However, it should be noted that our results are not directly comparable to those of previous systems, given that we trained the NLI model on a combination of the training data from the SNLI and MultiNLI corpora. Moreover, we did not optimize results for the NLI task, instead being more concerned with pre-training the network for building effective representations of textual inputs. It is perhaps the case that a more sophisticated matching method would improve the results for the NLI task~\cite{gong2017natural}, but in our pre-training scenario it is more important to have a good performance on the part of the network that encodes the semantics of the input sentences~\cite{supervised}.

\begin{table}[t]
  \begin{center}
    \caption{Results obtained over the testing split of the FNC-1 dataset.}
    \begin{tabular*}{\textwidth}{l @{\extracolsep{\fill}} c @{\extracolsep{\fill}} c @{\extracolsep{\fill}} c @{\extracolsep{\fill}} c @{\extracolsep{\fill}} c @{\extracolsep{\fill}} }
    
       & \textbf{Weighted} & \multicolumn{4}{c}{\textbf{Per-Class Accuracy} }\\
      \cline{3-6}
      \textbf{Method} & \textbf{Accuracy} & \textbf{Unrelated} & \textbf{Discuss} & \textbf{Agree} & \textbf{Disagree}\\
      \hline
       FNC-1 baseline\footnote{\scriptsize{\url{http://github.com/FakeNewsChallenge/fnc-1-baseline}}} & 75.20 & 97.97 & 79.65 & 9.09 & 1.00\\
       Baseline based on word2vec + hand-crafted features~\cite{combination} & 72.78 & 96.05 & 53.38 & 50.70 & \textbf{9.61}\\
       Baseline based on skip-thought embeddings~\cite{combination} & 76.18 & 91.18 & 81.20 & 31.80 & 0.00\\
       Baseline based on TF-IDF vectors~\cite{combination} & 81.72 & 97.90 & 81.38 & 44.04 & 6.60\\
       Best set of features from Tosik et al.~\cite{tosikdebunking} & 78.63 & 97.98 & \textbf{90.95} & 1.42 & 0.00\\
       \hline
       Neural baseline based on bi-directional LSTMs~\cite{combination} & 63.11 & 78.27 & 58.13 & 38.04 & 4.59\\
       Neural method from Mohtarami et al.~\cite{Mohtarami2018} & 78.97 & -- & -- & -- & --\\
       Neural method from Mohtarami et al. + TF-IDF~\cite{Mohtarami2018} & 81.23 & -- & -- & -- & --\\
       \hline
       3\textsuperscript{rd} place at FNC-1 -- Team UCL Machine Reading \cite{riedel2017simple} & 81.72 & 97.90 & 81.38 & 44.04 & 6.60\\
       2\textsuperscript{nd} place at FNC-1 -- Team Athene\footnote{\scriptsize{\url{http://medium.com/@andre134679/team-athene-on-the-fake-news-challenge-28a5cf5e017b}}} & 81.97 & \textbf{99.25} & 80.89 & 44.72 & 9.47\\
       1\textsuperscript{st} place at FNC-1 -- Team SOLAT in the SWEN\footnote{\scriptsize{\url{http://blog.talosintelligence.com/2017/06/talos-fake-news-challenge.html}}} & 82.02 & 98.70 & 76.18 & 58.50 & 1.86\\
      \hline
       Previous state-of-the-art -- Bhatt et al.~\cite{combination} & \textbf{83.08} & 98.04 & 85.68 & 43.82 & 6.31\\
       \hline
       MLP with the considered external features & 81.95 & 97.86 & 77.93 & 52.55 & 2.87\\
       \hline
       Bi-LSTM + max-pooling & 81.29 & 96.99 & 80.76 & 45.35 & 5.16\\
       Bi-GRU + max-pooling + attention & 80.76 & 97.36 & 76.08 & 52.08 & 4.16\\
       Bi-LSTM + max-pooling + attention {\it (best encoder)} & 82.23 & 96.74 & 81.52 & 51.34 & 10.33\\
       Stacked bi-LSTMs + shortcuts + max-pooling & 82.16 & 96.13 & 79.39 & 56.59 & 11.91\\
       Stacked bi-GRUs + shortcuts + max-pooling + attention & 81.95 & 96.22 & 77.71 & 52.86 & \textbf{19.08}\\
       Stacked bi-LSTMs + shortcuts + max-pooling + attention & 81.16 & 95.25 & 73.32 & \textbf{66.47} & 6.74\\
       \hline
    \end{tabular*}
  \end{center}
\end{table}

Table 3 presents results obtained over the FNC-1 testing dataset, including results from (a) previous baseline methods leveraging feature engineering, (b) baseline methods using simpler neural networks, (c) the best submissions to the FNC-1 challenge, (d) the previous state-of-the-art on the FNC-1 testing dataset, (e) a Multi-Layer Perceptron (MLP) leveraging the complete set of external features outlined in Section 3.4, and (f) the proposed neural architecture, considering all the components and model pre-training with the NLI datasets. Each group in Table 3 corresponds to one of the previous six items, with the last group presenting the results obtained with different sentence encoders, i.e. considering only one bi-RNN or a stack of two bi-RNNs with shortcut connections, and a combination of max-pooling and neural attention. The best weighted accuracy was achieved with a sentence encoder leveraging a single bi-LSTM, similarly to what was obtained in the case of the NLI datasets. The complete neural model, leveraging the best sentence encoder, outperformed the baseline method based on a MLP with the considered set of external features (which also achieved competitive results, outperforming the third best team at the FNC-1 competition and performing similarly to the more complex neural network architecture from Mohtarami et al.~\cite{Mohtarami2018}, in terms of the weighted accuracy metric).

We also evaluated variations of the complete method introduced in Section 3, using the best sentence encoder but without considering (a) model pre-training with the NLI datasets, (b) external features complementing the neural representations, (c) the matching between the headline and the first two sentences of the article, and (d) combinations of the three components above. Table 4 displays the results for this set of ablation tests.

Our full model without the branch matching the headline against the first two sentences of the body slightly outperforms the previous state-of-the-art in the FNC-1 test split, obtaining a weighted accuracy score of 83.38\%. The different ablation tests in which this branch has been removed seem to indicate that the matching against the first two sentences does not provide useful/additional information to the FNC-1 task. The results also show that both model pre-training and the external features are essential to achieving good results with the proposed neural network architecture, confirming previous studies that have shown that the FNC-1 task is challenging for methods based exclusively on representations built through deep neural networks. 

When the weights of the network are not initialized with the weights from the NLI task (i.e., when there is no model pre-training), the weighted accuracy drops to 81.85\% (or 82.02\% if the matching between the headline and the first two sentences is also removed). If the external features are instead removed, the weighted accuracy drops to 75.31\%, which is better than a baseline method leveraging a bi-directional LSTM~\cite{combination}, but only marginally better than the official baseline released by the organizers of FNC-1, and much worse than the winning entries at the challenge, or the previous state-of-the-art by Bhatt et al.~\cite{combination}.  
\begin{table}[t]
  \begin{center}
    \caption{Experimental results with different variations of the proposed method, over the testing split of the FNC-1 dataset.}
    \begin{tabular*}{\textwidth}{l @{\extracolsep{\fill}} c @{\extracolsep{\fill}} c @{\extracolsep{\fill}} c @{\extracolsep{\fill}} c @{\extracolsep{\fill}} c @{\extracolsep{\fill}} }
    
       & \textbf{Weighted} & \multicolumn{4}{c}{\textbf{Per-Class Accuracy} }\\
      \cline{3-6}
      \textbf{Method} & \textbf{Accuracy} & \textbf{Unrelated} & \textbf{Discuss} & \textbf{Agree} & \textbf{Disagree}\\
       \hline
       Best encoder with the complete architecture & 82.23 & 96.74 & \textbf{81.52} & 51.34 & 10.33\\
       \hline
       Best encoder - pre-training & 81.85 & \textbf{98.26} & 73.61 & 63.16 & 10.19\\
       Best encoder - matching first sentences & \textbf{83.38} & 97.27 & 80.60 & 59.64 & 13.06\\
       Best encoder - external features & 75.31 & 92.66 & 72.20 & 43.72 & 4.73\\
       Best encoder - pre-training - first sentences & 82.06 & 97.38 & 67.20 & \textbf{74.62} & \textbf{22.53}\\
       Best encoder - pre-training - first sentences - features & 76.08 & 86.23 & 72.65 & 61.90 & 15.49\\
       \hline
    \end{tabular*}
  \end{center}
\end{table}

\begin{figure}[t!]
  \begin{center}
  \includegraphics[width=0.475\textwidth]{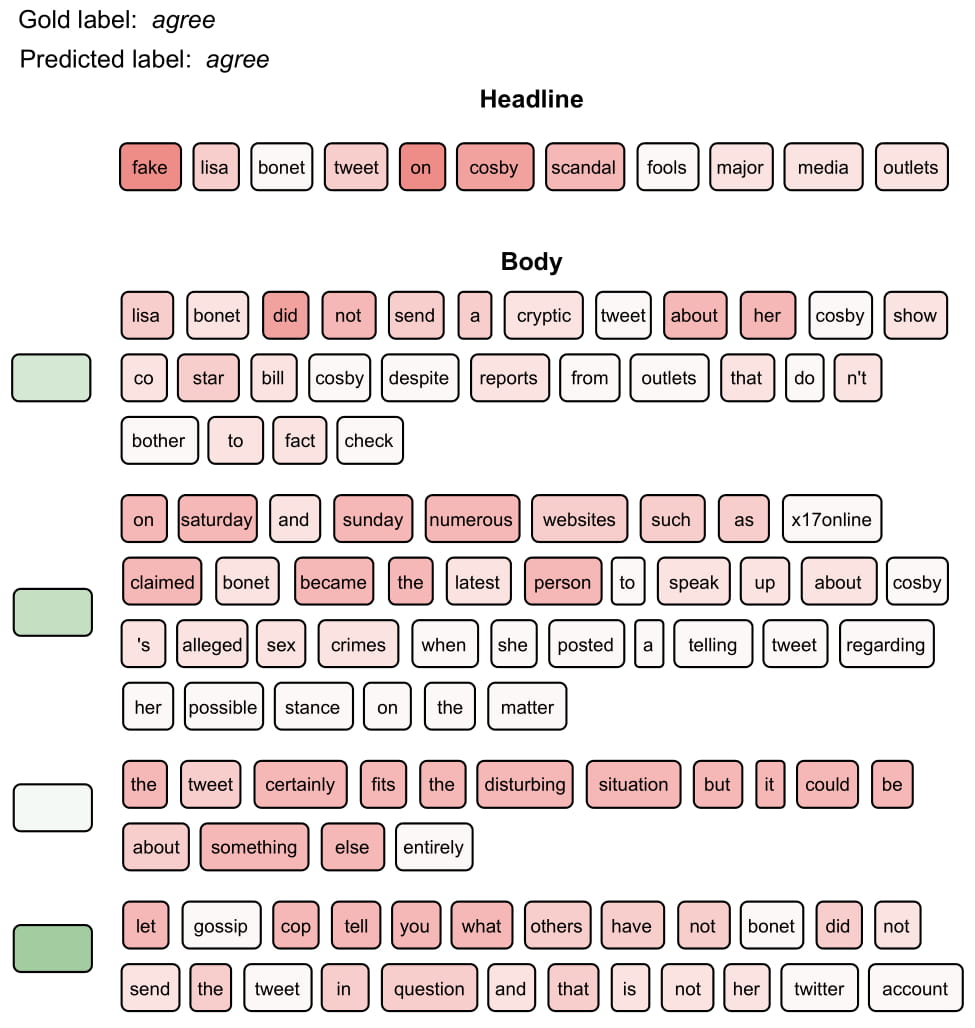} ~~~~~~~~~~~~~~~~~~ \includegraphics[trim=0 -3.9cm 0 0, width=0.475\textwidth]{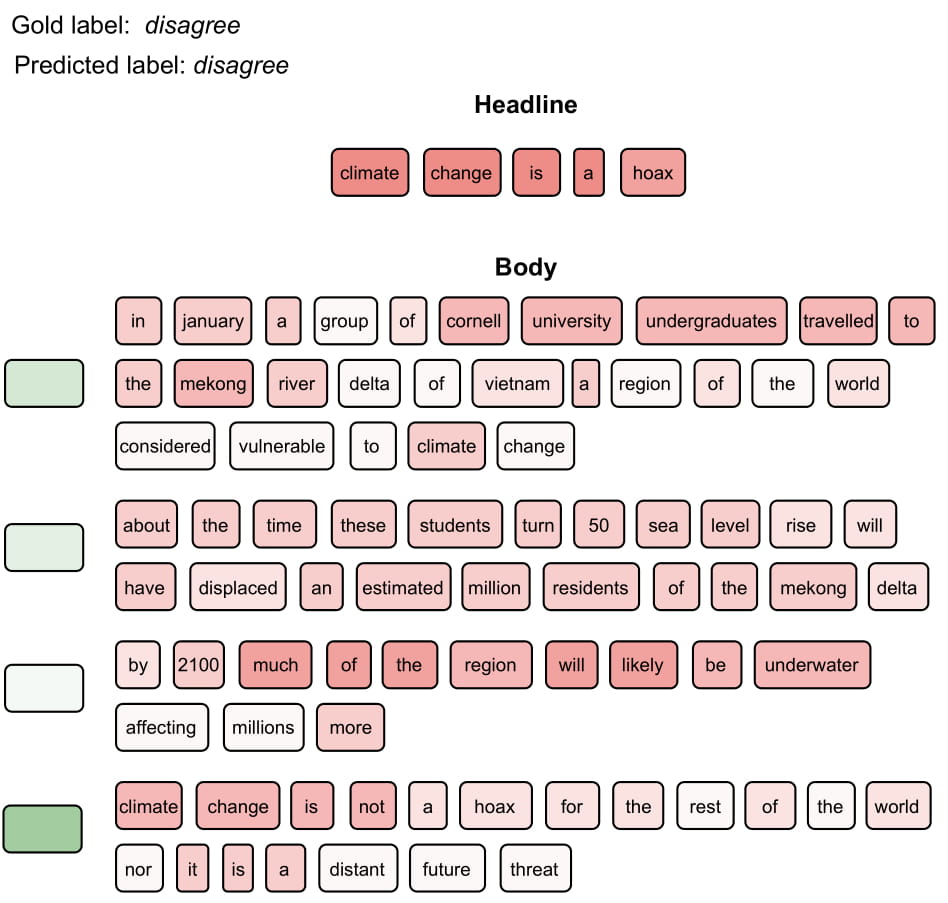}
  \caption{Attention weights computed with the complete model, for two example instances from the FNC-1 test split.}
  \label{fig:att1}
  \end{center}
\end{figure}


\begin{figure}[b!]
  \begin{center}
  \includegraphics[width=0.495\textwidth]{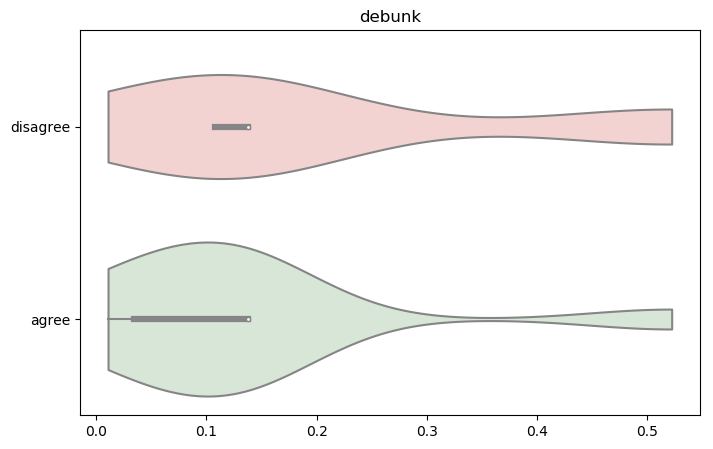}  
  \includegraphics[width=0.495\textwidth]{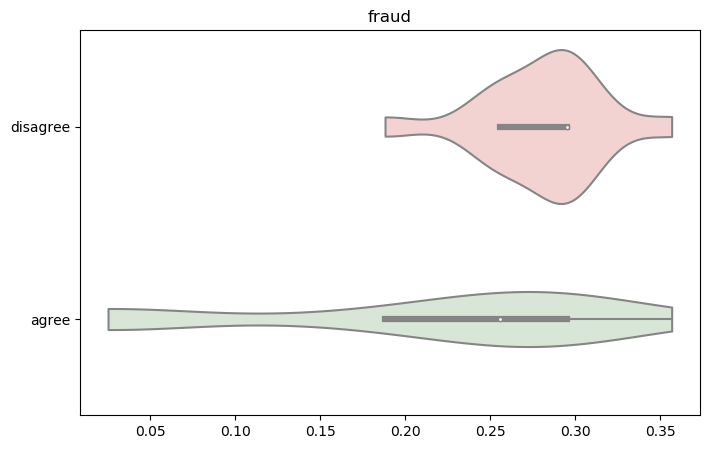} 
  \caption{Distribution for the neural attention weights associated to two distinct refutation words, within FCN-1 testing instances belonging to the {\it disagrees} versus the {\it agrees} class.}
  \label{fig:att1}
  \end{center}
\end{figure}

Besides evaluating model quality in terms of predictive accuracy, we also attempted to see if the neural attention weights associated to the sentence encoder, and to the hierarchical encoder for the news article body, could provide clues as to the rationale behind particular classification decisions. Figure 4 illustrates the attention weights computed by the neural network for two example instances from the FNC-1 testing dataset, with darker colors corresponding to the words/sentences to which the model assigned a higher weight. The two example instances shown in Figure 4 were correctly classified by our model (i.e., the best approach from Table 4), with one of them belonging to the \textit{agree} class and the other to the \textit{disagree} class. In both examples, the headline and the body of the news article contain words that can be highly indicative of the stance (e.g., words such as \textit{fake}, \textit{fraud} or \textit{hoax}, present in the lexicons used to compute the refutation and polarity features).

The headline for the \textit{agree} example refers to a fake tweet, with the word \textit{fake} having a high weight. Some of the sentences in the body confirmed that the supposed author did not send the tweet, with words/expressions such as {\it did not send} or {\it claimed} also having a high weight. On what regards the \textit{disagree} example, its headline simply states \textit{climate change is a hoax}, while the last sentence of the corresponding news article (i.e., the sentence with the highest weight) denies the claim from the headline, containing the words \textit{climate change is not a hoax}.

Figure 5 further illustrates the weights computed for the neural attention layers in association to specific words, specifically by presenting 2 violin plots with the distribution for the attention weights associated to the words {\it fraud} and {\it debunk} (i.e., words that belong to the FNC-1 refutation dictionary and that also appear frequently in the test instances). The plots contrast normalized values for the attention weights (i.e., we normalized the attention scores associated to words in the headline and in the body text, using the min-max normalization scheme together with the minimum and maximum attention weights that are seen in these two parts of the input) that are estimated by the classifier when analyzing test instances belonging to the {\it disagrees} versus the {\it agrees} category. As expected, the weights associated to aforementioned words tend to be higher in the case of documents belonging to the {\it disagrees} class, thus confirming that the attention layers are indeed providing interesting information that can support model interpretability and the design of user interfaces for manually inspecting the classification results.

In brief, the obtained results with the FNC-1 dataset confirm the importance of using external similarity features for tasks involving the modeling/matching of long pieces of text, and also that model pre-training can be a simple, yet effective, way of improving the representations build through deep neural networks.

\section{Conclusions and Future Work}

This article presented a deep learning method for addressing the stance detection problem from the Fake News Challenge (FNC-1), leveraging bi-directional RNNs together with max-pooling and neural attention mechanisms for building representations from headlines and from the body of news articles, and combining these representations with external similarity features. We also explored the use of external sources of information (e.g., large datasets proposed for the natural language inference task) in order to pre-train specific components of the neural network architecture (e.g., the RNNs used for encoding sentences).

The obtained results show that our model, particularly when considering pre-training and the combination of neural representations together with external similarity features, slightly outperforms the previous state-of-the-art. Most of the previous methods proposed for the FNC-1 task leveraged careful feature engineering, instead of representation learning through deep neural networks. Our results also confirm the challenges in addressing the FNC-1 task with modern neural approaches, given that the external features were essential to the good performance of our model (i.e., they increased the weighted accuracy score by 6.92 percentage points).

Our experiments also confirmed the effectiveness of model pre-training leveraging Natural Language Inference data, confirming previous results by Conneau et al.~\cite{supervised}. When pre-training the sentence encoder with the SNLI and MultiNLI datasets, our model reaches a new state-of-the-art result on the FNC-1 testing dataset.

Despite the interesting results, there are also many possible ideas for future work. For instance, within the context of NLI, previous studies have proposed more advanced methods for modeling sentences~\cite{choi2018cell,dr-bilstm,tay2017compare,tay2017costack}, and/or for matching representations for the premise and hypothesis sentences~\cite{gong2017natural,dr-bilstm,tay2017costack}. Examples include methods that leverage discourse markers (i.e., words used to link two clauses in a sentence) through transfer-learning approaches~\cite{dman}, methods for modeling the interactions between the sentence representations as a tensor, afterwards applying a CNN over this tensor~\cite{gong2017natural}, or methods that create representations for each sentence that depend on the other sentence, e.g. by initializing the RNN responsible for encoding a sentence with the last hidden state of the RNN responsible for encoding the other sentence~\cite{dr-bilstm,tay2017compare,tay2017costack}. For future work, we plan to experiment with these and other similar ideas for extending the deep neural network used in the FNC-1 task, leveraging more recent developments within the general context of natural language understanding (e.g., methods that, instead of using RNNs for encoding text, consider only feed-forward computations and attention approaches, similar to those from the transformer architecture~\cite{transformer,cer2018universal,Dehghani2018}).

It is also our belief that extensions of the method presented in this article can latter find application in the context of tools for fake news detection, going beyond predicting the stance~\cite{perez2017automatic,DeClarE2018,konstantinovskiy2018towards}. Moreover, the integration of text analysis methods into practical tools for fake news detection can benefit from mechanisms to help interpreting the predictions, and neural attention mechanisms can be explored in this direction~\cite{hierarchical}. For future work, we would like to extend the experiments related to the analysis of the attention weights given to specific parts of the headline and/or the body text, for instance considering sparsemax~\cite{martins2016softmax} as an alternative to the softmax normalization in the attention weights, in the hope of obtaining a more selective and compact attention focus, facilitating interpretability. The previous study by Mohtarami et al.~\cite{Mohtarami2018}, exploring neural mechanisms for FNC-1, also provided some ideas regarding model interpretability, interesting to explore as future work (i.e., the authors extracted snippets of evidence for the predictions, although not with basis on neural attention mechanisms).

Besides leveraging textual information and the stance of headlines towards longer pieces of text, fake news detection can also benefit from the combination of text mining methods, such as the one advanced in this article, with other types of approaches. For example, some previous studies have noted that fake news stories are often relatively simple (i.e., fake stories are usually \textit{flatter} than ordinary news stories), using exaggeration in a way that makes it easy to detect and follow them \cite{alternative}, or mentioning facts that can easily be disproven through inconsistencies against external sources of information\footnote{\scriptsize{\url{http://fakenews.publicdatalab.org/}}}. Fake news stories also have particular geographical diffusion patterns~\cite{alternative}, with rapid reproduction and elevated mutation rates (i.e., fake news rise and fall in weeks and often in days). Future work in the area can perhaps consider the combination/extension of text mining methods with other types of external features, capturing different characteristics of news articles.

\section*{Acknowledgements}

This research was supported by Funda\c{c}\~{a}o para a Ci\^{e}ncia e Tecnologia (FCT), through the project grants with references CMUPERI/TIC/0046/2014 (GoLocal) and POCI/01/0145/FEDER/031460 (DARGMINTS), as well as through the INESC-ID multi-annual funding from the PIDDAC programme, which has the reference UID/CEC/50021/2013. We also gratefully acknowledge the support of NVIDIA Corporation with the donation of the Titan Xp GPU used in the experiments reported on this article.

\bibliographystyle{ACM-Reference-Format}
\bibliography{sample-bibliography}

\end{document}